\documentclass[10pt,twocolumn,letterpaper]{article}

\usepackage{times}
\usepackage{epsfig}
\usepackage{graphicx}
\usepackage{amsmath}
\usepackage{amssymb}
\usepackage{subfigure}
\usepackage{float}
\usepackage{indentfirst}

\usepackage[pagebackref=true,breaklinks=true,letterpaper=true,colorlinks,bookmarks=false]{hyperref}

\begin{document}

\title{\ Feature Fusion Use Unsupervised Prior Knowledge to Let Small Object Represent}

\author{Tian Liu, Lichun Wang, Shaofan Wang\\
Beijing University of Technology\\
{\tt\small liutian126@emails.bjut.edu.cn}
}

\maketitle

\begin{abstract}

Fusing low level and high level features is a widely used strategy to provide details that might be missing during convolution and pooling. Different from previous works, we propose a new fusion mechanism called FillIn which takes advantage of prior knowledge described with superpixel segmentation. According to the prior knowledge, the FillIn chooses small region on low level feature map to fill into high level feature map. By using the proposed fusion mechanism, the low level features have equal channels for some tiny region as high level features, which makes the low level features have relatively independent power to decide final semantic label. We demonstrate the effectiveness of our model on PASCAL VOC 2012, it achieves competitive test result based on DeepLabv3+ backbone and visualizations of predictions prove our fusion can let small objects represent and low level features have potential for segmenting small objects.
\end{abstract}

\section{Introduction}\label{intro}

Semantic segmentation task\cite{tian2019decoders,zhang2018context_attention,deeplabv3+,ASPP} labeling image pixel-wisely is very popular and competitive. Deep networks have been approved successfully for semantic segmentation and the network goes deeper and deeper to produce reliable high level semantic features. For deeper net, some methods fuse multiple levels feature maps and combine receptive fields\cite{ ASPP,li2018pyramid_attention,chen2017deeplab} to resolve disappearance of small objects and boundary in which feature fusion plays a key role. Concatenating and adding are the most simple and widely used ways to fuse features from different levels. Attention mechanism\cite{vaswani2017attention,zhang2018context_attention,li2018pyramid_attention} is another complex and delicate way to combine features according to weights learned under the supervision of semantic labels. 

Concatenating fusion gives more trust for high level features, so the channels of low level feature map are usually 1/4 or 1/5 of high level feature map, which makes low level features being neglected even if they were fused. Both adding and attention fusion compute final features by assigning proportions of high level feature and low feature pixel-wisely to make the feature addition, which is smoothing gap between high level features and low level features, so it has no clear tendency for feature level. 


Xiao \textit{et at} \cite{xiao2018unified} used low level feature map solely to predict textures and material of complex scene, which proves low level features have definite semantic. On the other hand, considering task for segmenting small objects, simple convolution network\cite{fraction_pool}  achieved 96.33\% accuracy in 2014 on the Cifar-10, in which resolution of the image is $32*32$. So simple and shallow networks have enough ability to describe detailed features of images with small size, but the details will disappear on high level feature maps while using more convolution levels. 

In our work, the high level and low level features give independent decisions with help of priory information. Different from former ideas, low level feature can make decision all by its own on some small region. Thus small object and edge detail showing up on small region will not be diluted by high level features. 

\begin{figure}
    \centering
\subfigure[]{
\label{picture}
\centering
\includegraphics[width=2.5cm,height=2.5cm]{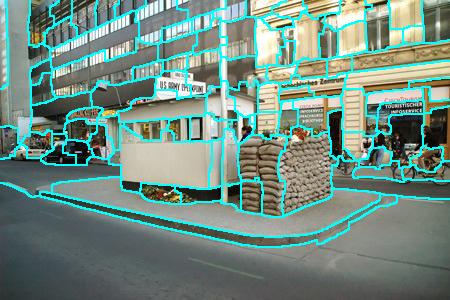}
\includegraphics[width=2.5cm,height=2.5cm]{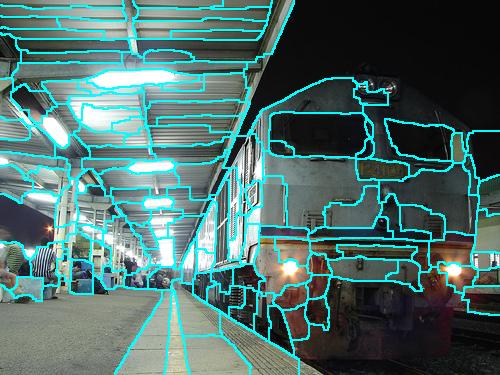}
\includegraphics[width=2.5cm,height=2.5cm]{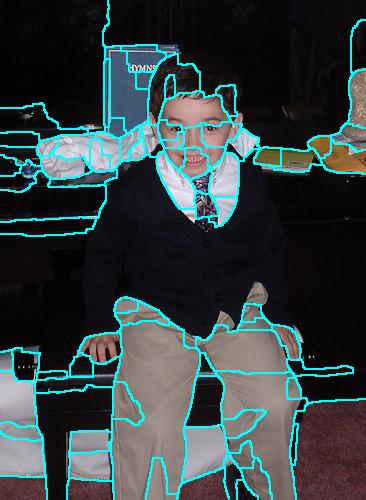}
}
\caption{\label{superexam}Examples of gPb/UCM over segment. }
\end{figure}

In order to make the region using different level features have more semantic independence, we take edge sensitive information distinct region adaptive to different level features. Notice that the region do not need to be exact objects, \textit{i.e.} superpixels generated by over segmenting images is a suitable choice as shown in Fig.~\ref{superexam}, since each superpixel describes a region having uniform semantic. Among all kinds of over segment methods, gPb/UCM~\cite{ucm} produces segmentation without considering the superpixels' size which focuses more on the semantic of region.

In summary, our major contributions include: 
\begin{itemize}
  \item [1)] 
 We propose a novel feature fusion method based on prior knowledge called FillIn, which can be used in any tasks that strengthen small objects.
 \item[2)]
 Low level features are used for independently deciding semantic labels of region such as small objects which may disappear during deeper convolution, and the predominant right of high level feature in prediction are remained at the same time.
 \item[3)] 
 Proportion of features from different level for generating final feature map can be controlled during fusion process.
\end{itemize}

\section{Related Work}
\begin{figure*}[]
\begin{center}
\includegraphics[width=.9\linewidth]{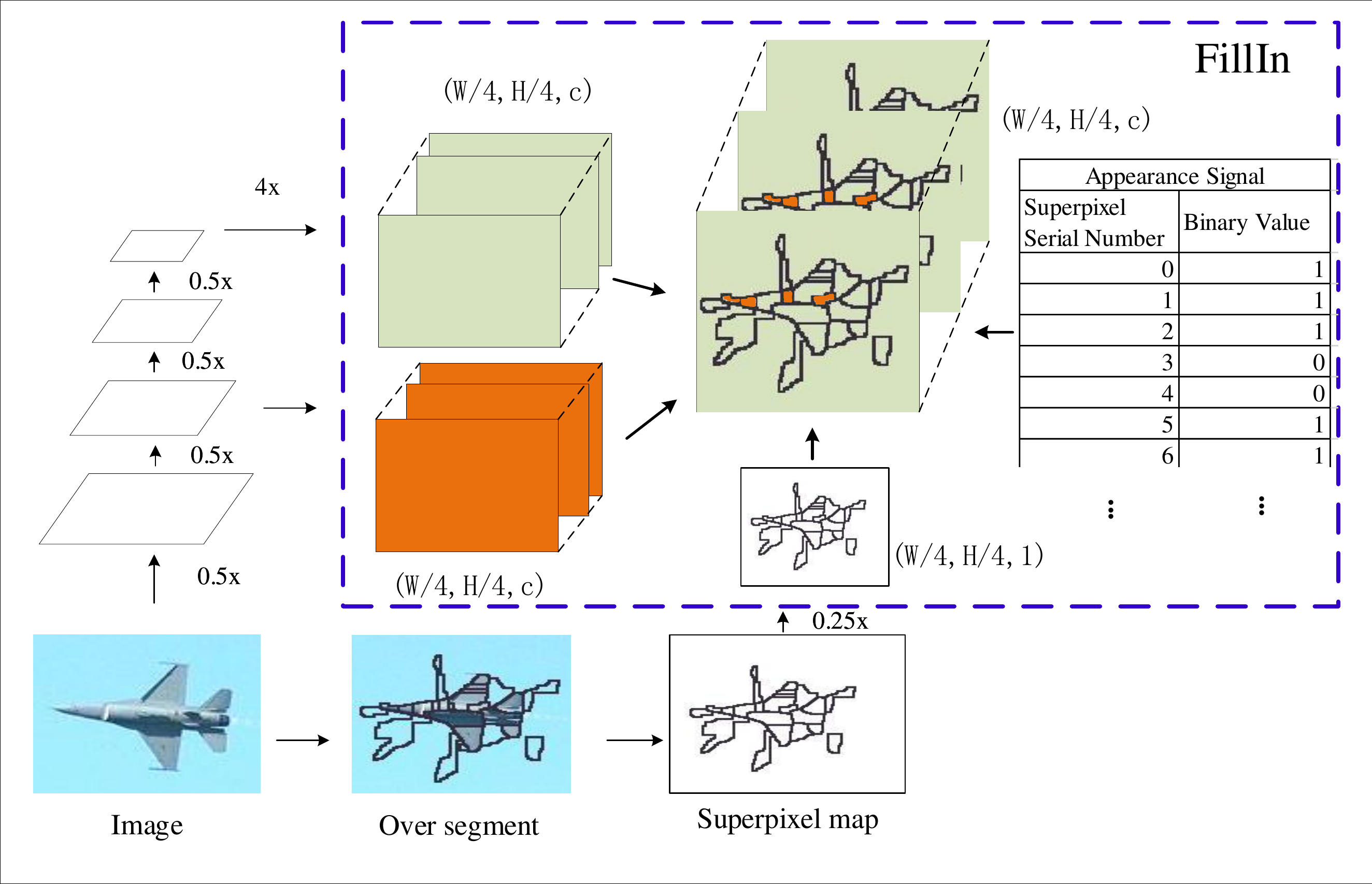}
\end{center}
   \caption{The flow chart of feature fusion. Our proposal have two inputs to the network: images and superpixel maps. FillIn combination in big blue square is our major innovation. The image need to be over segmented to get the superpixel maps which showed in black lines on the plane. FillIn take two different level features, higher layers in green and lower layers in orange, bilinear up-sampled to identical size if needed. We use superpixel map as a small object sensitive map downsampled and fed into FillIn. The forth input is Appearance Signal(AS) generated in Sec.~\ref{filter} which provide binary value for every superpixel pieces of original size, so smaller size superpixel map can use it too.
   The process can be a metaphor to children coloring drawing. high and low level feature map are pigments, and superpixel map is the image with only contours, and Appearance Signal is the guidance telling which part should painted in what color. In our method, for each slice, fill the region on new feature map with high level feature if its according region's serial number on sensitive map projected to binary value 1, else fill it with low level feature map. Better viewed in color.}
\label{fig:flow}
\end{figure*}

For feature fusion, there are two effective and simple ways to fuse feature maps.

Adding fusion is widely applied in shortcut connections, by which Resnet\cite{he2016resnet} relives the gradient vanishing problem and makes training deeper network possible. Adding is also used in some attention\cite{li2018pyramid_attention} networks to combine features of different layers. Dissimilar from directly adding two feature maps, attention mechanism\cite{zhang2018context_attention} uses weights learned during network training to guide the pixel-wise fusion, which is proved to be effective to utilize low level features.

Concatenate is another simple fusion method\cite{deeplabv3+,chen2017deeplab,ASPP,tian2019decoders,chen2019hybrid}. And it is the major way to fuse feature maps of different semantic level before attention mechanism appears, since adding with low level feature map will dilute the features provided by higher layers. Normally, higher-layer feature contains more abstract semantic meaning and lower-layer contains more details. Concatenate can take advantage of the difference and complementation, for example,  Zhao \textit{et al} \cite{ASPP} suggests Atrous Spacial Pyramid Pooling(ASPP) utilizing atrous convolution\cite{chen2017deeplab} to get better semantic feature map from multi receptive field, which shows remarkable improvements. 

In fact, small objects sometimes can be expressed successfully due to feature fusion while the high level feature are given partial decision-making power\cite{ASPP,xception,tian2019decoders} at every pixels. But prediction for small object is always not well because the high level features make prominent decision for final semantic label. While concatenating, higher-layer feature maps usually contribute more channels\cite{deeplabv3+}. As to attention mechanism\cite{zhang2018context_attention}, ground-truth and high level global feature are usually used as guidance for feature fusion, so low level features have weak effect on deciding semantic label. Both the above strategies consider small object by fusing low level features, but the concerns are not enough for small object labeling. The decision is actually made by big objects which hold plenty information on high level feature map. Then fusing features are then average of high level and low level features on pixel-wise\cite{li2018pyramid_attention,chen2017deeplab}, which fails to guarantee expression of lower levels.

Different from previous methods, we propose a feature fusion method that allocates region to different level features based on prior knowledge, so that final semantic prediction of some small region is made by low level feature alone.

\section{Our method}

In this section, we will present the FillIn combination shown in Fig.~\ref{fig:flow} and generation of Appearence Signal(AS). We take DeepLabv3+ \cite{deeplabv3+} as backbone to illustrate using different Filter scale in four network structure.

\subsection{FillIn combination}
\label{fillin}

Our purpose is to protect the low level features to be independent from the high level features, so that some small objects can be labeled by low level features.So we need a small objects sensitive map which can catch suspicious patches thus provide map segmented in region for features of different level to fill in. Notice that we do not need accurate boundary, because accuracy of segmentation would be provided by feature maps. Since Fillin is exclusive operation: a region can only be filled by either higher layers or lower layers, the small objects fill in small region is able to to remain in feature fusion.

More specifically, the sensitive map is a superpixel map which is generated through over segment images, but traditional superpixels standard demand superpixels pieces better in equal size. So we use gPb/UCM, because it over segments images regardless of the magnitude. We also need  guidance to indicate which region have proper sizes for feature maps of different levels. So we simulate down sampling to sift small pieces for small objects in Sec~.\ref{filter}.

In Fig.~\ref{fig:flow} we demonstrate the FillIn method in decode scale $4, [W/4,H/4,C]$, and use metaphor to children coloring drawing to help understand and illustrate that our method and idea is very simple, and  better viewed in color. We would also give a formulaic description. 
Superpixel map is actually a integer matrix, $\mathbf{S}$, and $\mathbf{S} \in \mathbb{R}^{m \times n}$. The Appearance Signal function defined in Eq.~\eqref{as} will indicate which superpixel pieces $x$ remains in high level can be formulated as

$$\mathbb{I}(x)$$

$\mathbb{I}(x)$ is the indicator function which returns 1 if the predicate x is true, and 0 otherwise. Downsample superpixel maps to the same size of feature map to be fused and the downsampling stride is $t$. In Fig.~\ref{fig:flow} $t$ is $4$ and denote as $0.25x$.

Formally, downsampling can be formulate in Eq.~\eqref{up}.
\begin{align}\label{up}
\mathbf{U}_{i,j}=\mathbf{S}_{min(i\times t,m),min(j\times t,n)}
\end{align}
, and generate a small superpixel map $\mathbf{U}$. We can then generate two binary matrix $\mathbf{H}$ and $\mathbf{L}$ 
$$\mathbf{H} = \mathbb{I}(\mathbf{U}) \\$$, and $$ \mathbf{L} = \mathbf{1}-\mathbb{I}(\mathbf{U}) $$ where $\mathbb{I}(\cdot)$ is the matrix form of $\mathbb{I}(x)$

Finally our fused feature map can be formulated in
$\mathbf{F}\in\mathbb{R}^{m\times n\times C}  $ 
$$\mathbf{F}_{:,:,c}^{fused} = \mathbf{F}_{:,:,c}^L \odot \mathbf{L}+ \mathbf{F}_{:,:,c}^H \odot \mathbf{H}$$
where $\odot$ is the element-wise product, and $c=1,2,\cdots,C$.

Bilinear upsample is used to generate equal feature map,identical in depth, width and height in our work. Since we use same superpixel map on all slice of the new feature map, so the features along its depth will belong to the same level. 


\subsection{Appearance Signal}
\label{filter}
\begin{figure}
\begin{center}

\includegraphics[width=.78\linewidth]{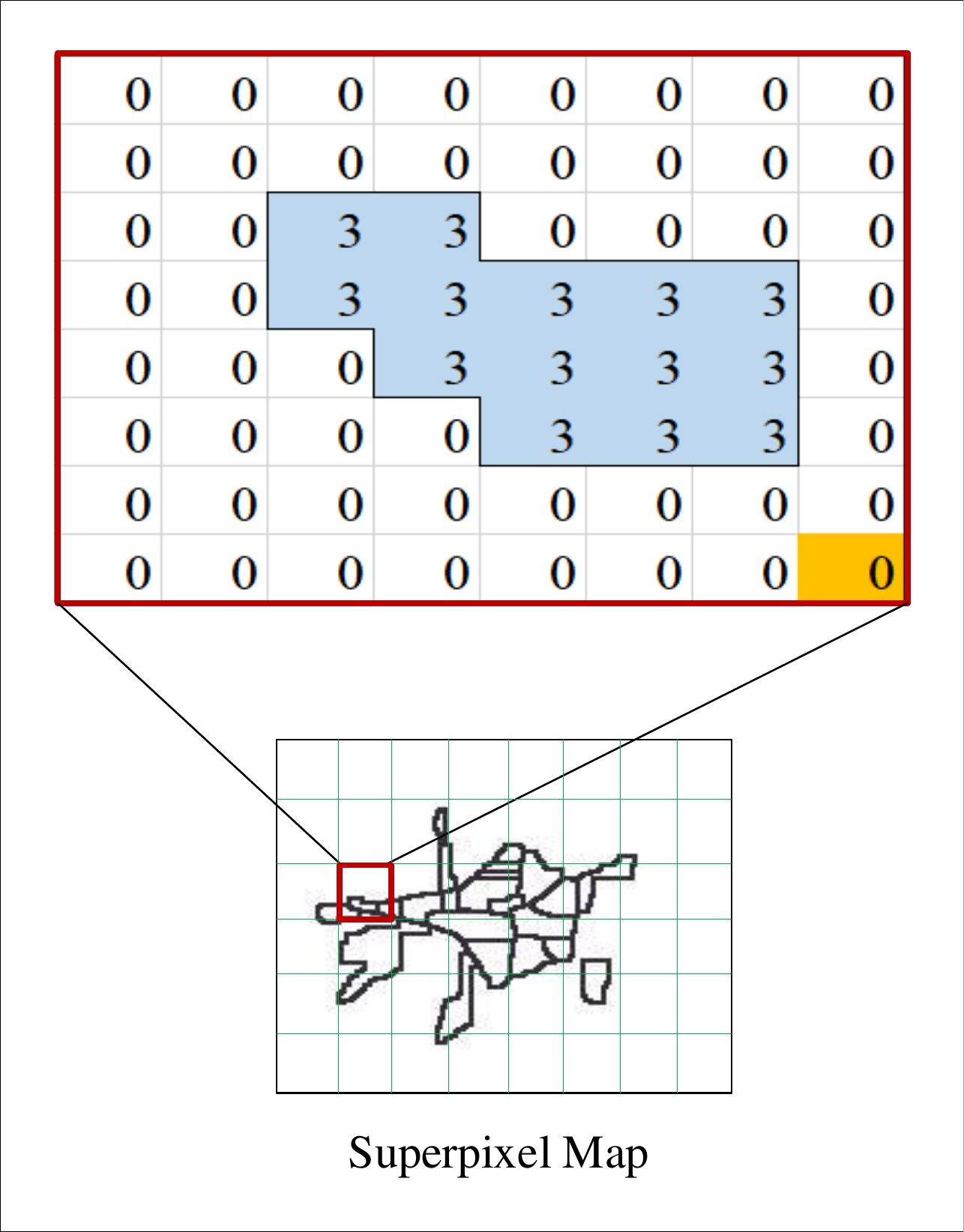}
\end{center}
   \caption{We show the superpixel pieces missing during downsampling to simulate small objects disappearing in a high level feature map. Downsample at the plane's head in small red box in this figure. Take stride 8 as an example. During down sampling , some details may be lost, so we downsample superpixel maps to catch pieces that are more likely to vanish on high level. As shown in big red box, the superpixels of the plane's head are represented by its sequential number 3, and the background is 0(To illustrate more clearly, we posit that the plane head is surrounded by background). Noticing superpixel piece 3 is fully covered by downsample box, if it  not picked then it will definitely lose during downsample. So, when down sampling the plane head, 0 is picked ,superpixel piece 3 will disappear and the according feature at the region of low level feature map might likely  vanish too.}
\label{fig:filter}
\end{figure}

Appearance Signal is a fair indication to guarantee the representation of small objects and remain the dominant right of big continuous semantic region. It is crucial that Appearance Signal come from unsupervised method and can leave small pieces to tiny object fairly. 

As we mentioned in Sec.~\ref{intro} that some semantic pieces might disappear during downsample and convolution, so we simulate downsample processes to cast the serial numbers of superpixels to binary values which indicate if that superpixel might disappear at high level. 
superpixel pieces that vanish after downsampling will be projected to $0$, and others will be $1$. 

more specifically, superpixel map of original size $\mathbf{S}$ have serial number set $V_s = \{x\mid x=\mathbf{S}_{i,j}\}$ 
We used the upsample in Eq.~\eqref{up} to get a smaller superpixel map $\mathbf{Q}$ under Appearance Signal(AS) stride $p$. Notice that the AS stride is independent from the network structure, and control the proportion of low level feature. Superpixel map $\mathbf{Q}$ have serial number set $V_q$ from $V_q=\{x \mid x=\mathbf{Q}_{i,j}\}$
Hence, $\mathbb{I}(x)$ will be define in Eq.~\eqref{as}.

\begin{align}\label{as}
\mathbb{I}(x)=\left\{\begin{matrix}
0 & x \in V_s - V_q\\
1 & x \in V_s \bigcap V_q
\end{matrix}\right.
\end{align}
where $x \in V_s$

As a result, pieces contain a square at size more that $[p+1,p+1]$ will not disappear. The whole superpixel piece is covered by some mutually exclusive downsample box and none of them picked serial number of this piece. And we used Fig.~\ref{fig:filter} to illustrate the on one condition when a region is discard. To demonstrate clearly we posit the head of plane $3$ is surround by another piece $0$. When the plane head is included in the red downsample box, and is not  picked, it will disappear in the high level feature map.

When the AP stride is bigger, the proportion of small region left to low level feature maps will increase. So AP stride depends on the reliability of your low level feature extractor. As Chen \textit{et al} in DeepLabv3+ \cite{deeplabv3+} use Xception\cite{xception} as backbone, We adopt AS stride 16.


\subsection{Network structure}

We use four decode structure to compare and shows our effect. All network have same encoder module and ASPP as DeepLabv3+, but decoder module is different.

\begin{figure}
\begin{center}
\includegraphics[ width=0.78\linewidth]{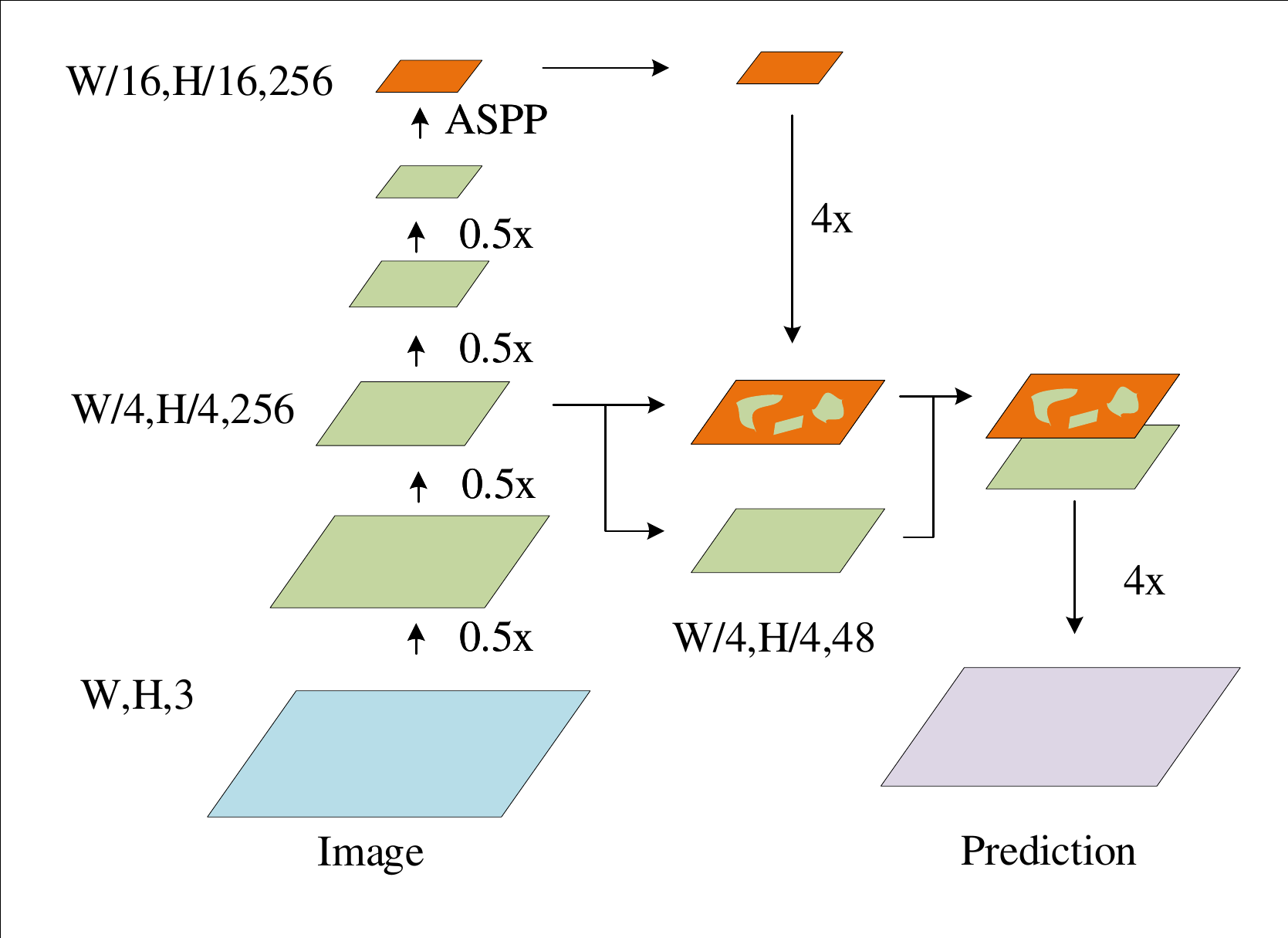}
\end{center}
   \caption{Bi4 Fillin scale is $4$. The mixed color of green and orange parallelogram is Fillin combiantion feature map which will take two $[W/4,H/4,256]$ features as input. Same as DeepLabv3+, we use conv low level feature to $48$ channels and concatenate it with FillIn fused feature map. Better viewed in color. Except for some region in according high level features replaced, all operations are same as DeepLabv3+\cite{deeplabv3+}}
   \label{fig:bi4}
\end{figure}

\textbf{Bi4: } As shown in Fig.~\ref{fig:bi4}, Only nuance exist in decoder module: The high level feature map demanded by concatenate is FillIn first with lower layer feature. As we mentioned above, we use bilinear upsample on low level feature map to FillIn with high level feature map.
To be clear, FillIn has no parameter and do not change feature shape. So the only dissimilar between the two models is that the some patch of high level feature is replaced by low level. 

\textbf{Bi2: }Difference between \textbf{Bi2} and \textbf{Bi4} is that \textbf{Bi2} has FillIn scale $2$, So we upsample two level of features to $[W/2,H/2,256]$first, then Fillin them. As for concatenate, we Conv low level feature to $[W/4,H/4,48]$ first, same as DeepLabv3+, and then bilinear upsample to the same size, $[W/2,H/2,48]$, as FillIn feature map. 

\textbf{Bi4ref:} \textbf{Bi4ref} has same FillIn structure as \textbf{Bi4}. But after the feature was combined and concatenated to $[W/4,H/4,304]$, we bilinear upsample it to scale $[W/2,H/2,304]$ and then Conv to predict .

\begin{figure}

\begin{center}
\includegraphics[ width=0.82\linewidth]{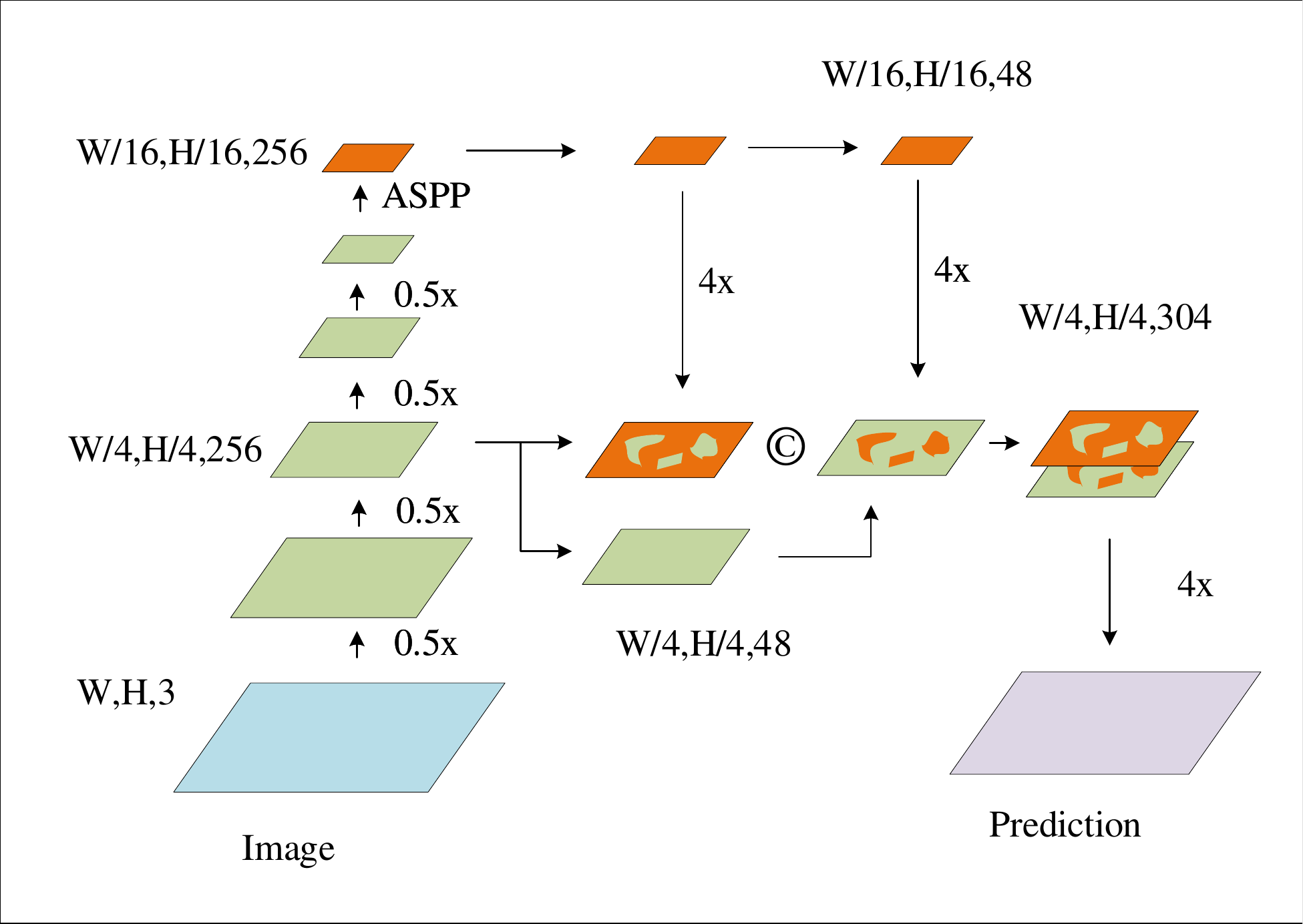}
\end{center}
   \caption{We concatenate two feature maps which have already fused by FillIn separately.}
\label{fig:revers}
\end{figure}

\textbf{Reverse: } 
Different from above structure. The thin feature in feature fusion can be FillIn feature map too, shown in Fig.~\ref{fig:revers}. The superpixel map is same in both thick and thin feature map but the Appearance signal is reverse. We use $$\mathbb{I}_{thin}(x)=\mathbf{1}-\mathbb{I}_{thick}(x)$$ where $\mathbf{1}$ is a matrix fill with 1.$\mathbb{I}_{thin}(x)$ is the indicator function of thin feature map fusion. $\mathbb{I}_{thick}(x)$ for thick feature map. let higher layer feature be the semantic supplement for lower layer feature map and keep low level feature map as a refinement to high level, which is a fair strategy for both small and big objects.

\section{Experiment evaluation}


We evaluate our model on PASCAL VOC 2012 semantic segmentation benchmark
which consist of $1,464$($train$), $1,449(val)$ and $1456(test)$ images. It have one background class and twenty foreground object classes. 

We build our work on DeepLabv3+ and TensorFlow. We use the pretrained SBD weight provide by Chen \textit{et al} \cite{deeplabv3+}
SBD pretraied weight is trained on COCO~\cite{coco}, ImageNet-1K\cite{imagenet}, SBD~\cite{SBD}dataset.
All of the hyperparameters is same as Chen\cite{deeplabv3+} train DeepLabv3+ for PASCAL VOC 2012$test$ sets\cite{voc}. We use batch size 24 and feeze BN parameter based on SBD pretrained weight, and freezing is also identical to Chen \cite{deeplabv3+}.
When training the batch normalization\cite{batch_normalization} parameter our result on SBD is lower than DeepLabv3+ for $2\%$. The SBD weight provided by DeepLabv3+ generate miou 82.2\%.

\subsection{Appearance Signal stride }

Filter stride decide the proportion of features in two level,more specifically, the bigger the stride is, and the more high level features will be replaced by low level. We compare Appearance(AS) stride on \textbf{Bi4ref} in Tab.~\ref{tab:stridetab}.
 The result train and eval under output stride 16\cite{deeplabv3+}, with multi scale and flip.
Notice that train on AS stride$=16$ achieve better result.


\begin{table}[]
\centering
\begin{tabular}{llll}
\hline
                & 8     & 16    & 24    \\ \hline
\textbf{Bi4ref} & 85.97 & \textbf{86.12} & 85.30 \\ \hline
\end{tabular}
\caption{Results of AS stride on PASCAL VOC 2012 $val$ setss.}
\label{tab:stridetab}
\end{table}

\subsection{FillIn scale and network structure}

Results of four networks shows Tab.~\ref{tab:fill scale tab}. We use same Filter stride and hyper parameter on this experiment. Simple results means evaluation only with output stride 16\cite{deeplabv3+} and complex results evaluate under output stride 8 and using multi scale input and flip. \textbf{Bi2} higher than \textbf{Bi4} for 0.02\% shows that using different FillIn scale have little effect with DeepLabv3+ backbone.As Chen  mentioned in their paper\cite{deeplabv3+}, the complex decoder has insignificant effect.

\textbf{Reverse} is slightly better than other work, might because the high level feature supply lower layers some global information.
\textbf{Bi4ref} has same convolution layers as \textbf{Bi4} and \textbf{Bi2}, and consider that we use pretrained SBD weight on DeepLabv3+, \textbf{Bi4ref} gain some benefits by remaining same combination scale and convolution on higher level.

\begin{table}[]
\centering
\begin{tabular}{lll}
\hline
                 & simple & complex \\ \hline
\textbf{Bi4}     & 84.04  & 86.04   \\ \hline
\textbf{Bi2}     & 83.83  & 86.07   \\ \hline
\textbf{Bi4ref}  & 83.81  & 86.12   \\ \hline
\textbf{Reverse} & \textbf{83.83}  & \textbf{86.19}   \\ \hline
\end{tabular}
\caption{Results of four structure on PASCAL VOC 2012 $val$ sets}
\label{tab:fill scale tab}
\end{table}

\subsection{State of art}

\begin{figure*}
    \centering
\subfigure[]{
\begin{minipage}[t]{0.23\linewidth}
\centering
\includegraphics[width=3.8cm,height=3cm]{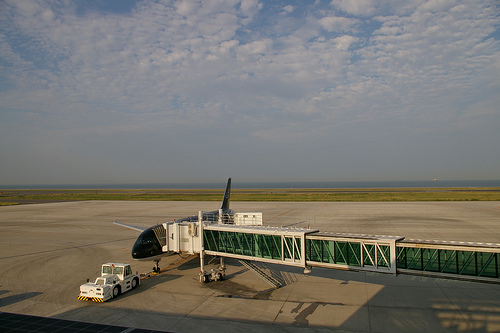}
\includegraphics[width=3.8cm,height=3cm]{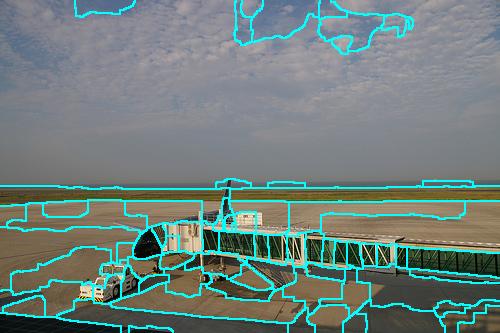}
\includegraphics[width=3.8cm,height=3cm]{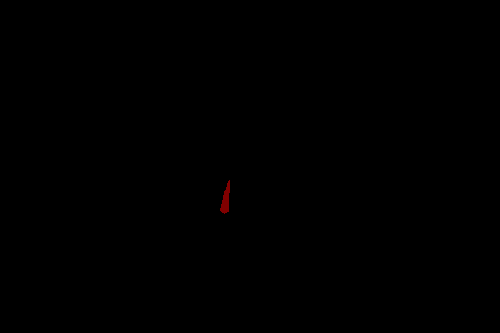}
\includegraphics[width=3.8cm,height=3cm]{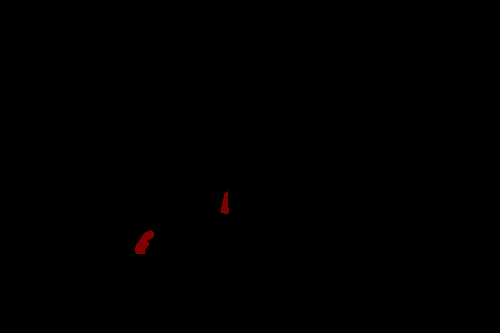}
\end{minipage}%
}
\subfigure[]{
\label{picture}
\begin{minipage}[t]{0.23\linewidth}
\centering
\includegraphics[width=3.8cm,height=3cm]{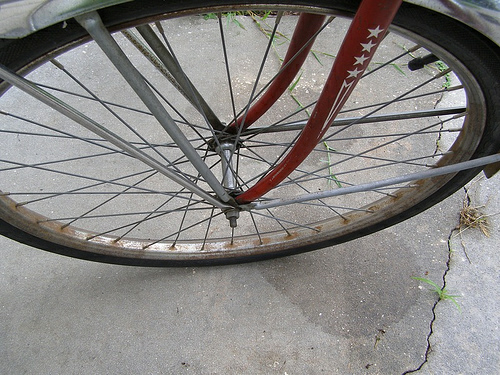}
\includegraphics[width=3.8cm,height=3cm]{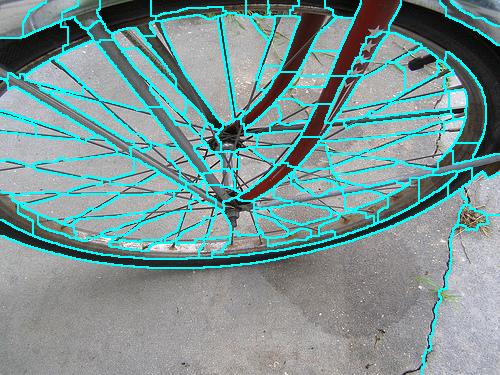}
\includegraphics[width=3.8cm,height=3cm]{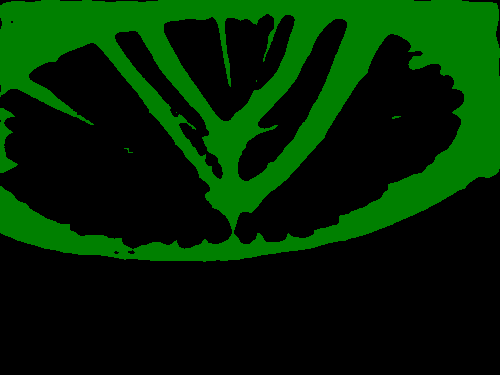}
\includegraphics[width=3.8cm,height=3cm]{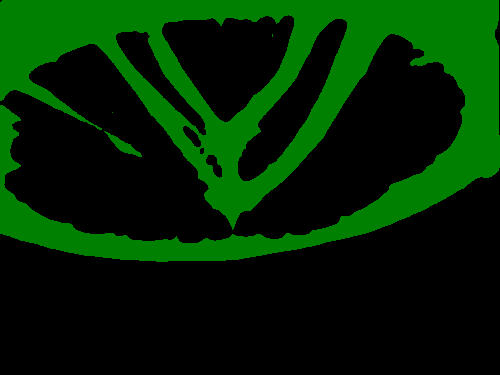}
\end{minipage}%
}
\subfigure[]{
\label{picture}
\begin{minipage}[t]{0.23\linewidth}
\centering
\includegraphics[width=3.8cm,height=3cm]{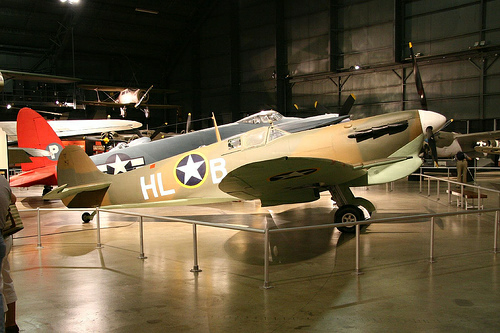}
\includegraphics[width=3.8cm,height=3cm]{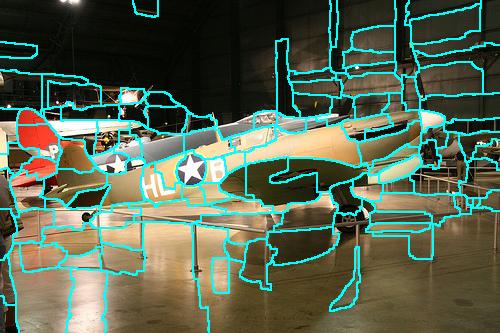}
\includegraphics[width=3.8cm,height=3cm]{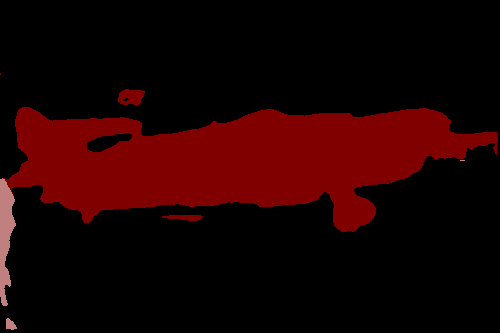}
\includegraphics[width=3.8cm,height=3cm]{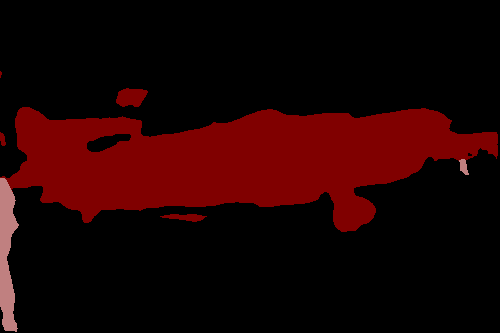}
\end{minipage}%
}
\subfigure[]{
\label{picture}
\begin{minipage}[t]{0.23\linewidth}
\centering
\includegraphics[width=3.8cm,height=3cm]{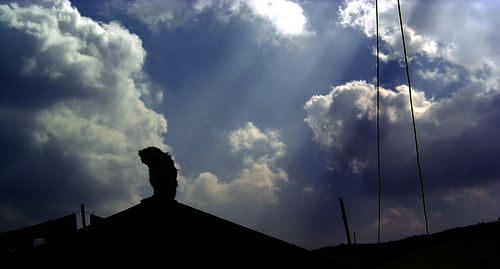}
\includegraphics[width=3.8cm,height=3cm]{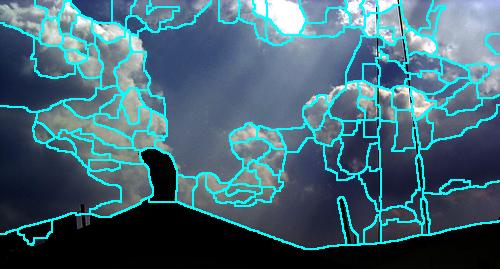}
\includegraphics[width=3.8cm,height=3cm]{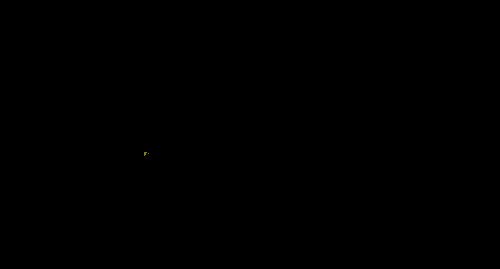}
\includegraphics[width=3.8cm,height=3cm]{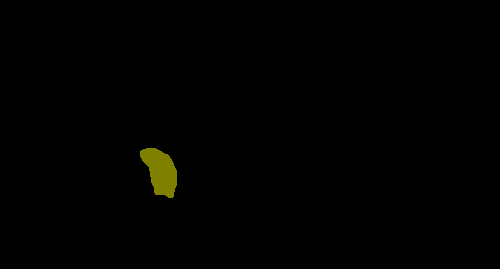}

\end{minipage}%
}


\caption{Visualization of our method(87.4\%) and DeepLabv3+(87.8\%) on PASCAL VOC 2012 test set, We use $output\ stride=8$, $batch\ size = 16$ and $AS\ stride=16$ on network \textbf{Bi4} . From top to bottom: images, superpixel maps, predictions of DeepLabv3+ and ours predictions. }
\label{fig:visual}
\end{figure*}

We achieve 87.9\% at PASCAL VOC 2012 test set \footnote{\url{http://host.robots.ox.ac.uk:8080/anonymous/YE7MNH.html}}shown in Tab.~\ref{tab:art}

\begin{table}[]
\centering
\begin{tabular}{ll}
\hline
              & test          \\ \hline
DeepLabv3+    & 87.8          \\ \hline
\textbf{ours(Bi4ref)} & \textbf{87.9} \\ \hline
\end{tabular}
\caption{Results of PASCALVOC 2012$test$ sets}
\label{tab:art}
\end{table}

We train on the $trainval$ sets based on pretrained SBD weight generated by DeepLabv3+ and freeze the BN parameters same as Chen \textit{et al} \cite{deeplabv3+}. Notice that we didn't achieve state of art result and Tian \textit{et al}  \cite{tian2019decoders} achieve 88.1\% with their decoder. But our feature fusion strategy provides competitive results and catches more small objects compared with DeepLabv3+ as shown in Fig.~\ref{fig:visual}. 

We use \textbf{Bi4ref} and train on $AS\ stride= 16$.
All evaluation is based on output stride 8, but we train on output stride=16 and DeepLabv3+ train on output stride 8. We used as much batch size as  we can which is 24. All other hyper parameters are same as Chen\cite{deeplabv3+} adopt in on PASCAL VOC $test$ sets.

As shown in Fig.~\ref{fig:visual}, clearly, low level features have the ability to predict far and small objects, complex and variation objects, seriously occluded objects, and slim part of objects
%

\section{Conclusion}

We have proposed a simple feature fusion method which utilize unsupervised prior knowledge as guidance, and it allows the low level feature maps predict small objects in tiny region. The fusion can protect small objects and remain the predominant semantic feature of higher layers.  Low level feature maps have the ability to predict small objects that are not able to be noticed by high level feature maps. But our method is not end-to-end and cost extra calculation on over segments, and utilize supervised signal for small object might helpful to the segmentation result in the future.

{\small
\bibliographystyle{ieee_fullname}
\bibliography{egbib}
}
\end{document}